# Initial Results on the Flora2 to OWL Bi-directional Translation on a Tabled Prolog Engine

**Abstract.** In this paper, we show our results on the bi-directional data exchange between the F-logic language supported by the Flora2 system and the OWL language. Most of the TBox and ABox axioms are translated preserving the semantics between the two representations, such as: proper inclusion, individual definition, functional properties, while some axioms and restrictions require a change in the semantics, such as: numbered and qualified cardinality restrictions. For the second case, we translate the OWL definite style inference rules into F-logic style constraints. We also describe a set of reasoning examples using the above translation, including the reasoning in Flora2 of a variety of ABox queries.

## 1 Introduction

Tremendous research has been invested in the last ten years into the Semantic Web services and logical formalisms to support discovery, contracting and execution of Web services. The current winning formalisms are, definitely, based on description logics, and, more specific, the Web Ontology Language (OWL)-DL language has been used to build several databases. Unfortunately, several problems have been discovered with this approach, such as: support for rules and logical updates. To solve these problems, we turned to logic programming, F-logic and Transaction Logic, as helping formalisms that support rules, frames and updates. In this paper, we describe our translation of knowledge bases from OWL representation language to the F-logic representation supported by the Flora2 system. Cyclic queries and terminologies can also be solved by our translation by using tabled top-down evaluation (available in [2] and [3]). In most cases, using this translation, we can reason both in the closed world assumption, as in logic programming, and in the open world assumption, as in Description Logics. Numbered and qualified cardinality restrictions can also be expressed with the Flora2 translation, but their semantics is different than the OWL semantics. Some OWL constructs cannot be expressed in our translation, such as: certain cases of disjunctions. We will describe these special cases in this paper and will detect and signal these cases in our translator. Our translation also extends the expressivity of OWL with constraints and role constructors: role intersection, composition, transitive closure, and inversion, these being easy to construct and reason with tabled Flora2 predicates.

The translation into Flora2's format makes possible the evaluation of transactions over the data in the ontology, making possible the design and execution of workflows and execution of plans that change facts about individuals while executing Web workflows. These features cannot be represented with the auto-epistemic K-operator and the reasoning tasks cannot be solved using the tableau algorithms (see updates of the ABox in DL-Lite in [4] and representation of supply chains in [5]).

The paper is organized as follows. The basic translations are defined in Section 2. Section 3 describes applications of the translation into querying and checking the integrity of ontology, and related work. Section 4 summarizes our contributions and concludes the paper.

## 2 Basic translations

This section presents the basic translations between the OWL description logic and the Flora2's F-logic language. A couple of simple operations are executed before any translation takes place: the short namespace URIs are replaced with full IRI (for example, "owl\#ontology" is replaced with "http://www.w3.org/2002/07/owl\#ontology"), XML types are replaced with the Flora2 types (for example, "xml\#string" is substituted with "_string").

We start by presenting the translation from OWL to Flora2. Some OWL syntax constructs have two translations into Flora2 F-logic: a deductive inference rule and a constraint. For instance, the "owl:allValuesFrom" restriction is translated into a constraint on all the values of a property from a specific class, but it is also used as a deductive inference rule, establishing that all values of that property are instances of a specific class. For instance, the OWL definition:

```
<owl:Class rdf:ID="Wine">
  <rdfs:subClassOf>
    <owl:Restriction>
        <owl:onProperty rdf:resource="#hasMaker"/>
        <owl:allValuesFrom rdf:resource="#Winery"/>
    </owl:Restriction>
  </rdfs:subClassOf>
</owl:Class>
```

is translated into the F-logic constraint:

```
Wine::_object[hasMaker *=> Winery].
```

and the inference rule:

```
?Y:Winery :- ?X:Wine, ?X:[hasMaker -> ?Y].
```

The two translations are complementary, that is, combined, they have the same semantics as the OWL construct.

We describe the translations from OWL to Flora2 F-logic in the following table.

**Table 1.** Mapping from OWL to Flora2 F-logic

| **OWL element** | **OWL example** | **OWL example translation** |
|---|---|---|
| rdfs:subClassOf | <owl:Class rdf:ID="Wine"><br><rdfs:subClassOf rdf:resource="&food;PotableLiquid" /><br></owl:Class> | Wine::'&food;PotableLiquid'. |
| owl:equivalentClass | <owl:Class rdf:ID="Wine"><br><owl:equivalentClass rdf:resource="Vin "/><br></owl:Class> | Wine :=: Vin<br>User-defined equality:<br>- all objects of one class are also objects of the second class:<br>?X:C1 :- ?X:C2.<br>?X:C2 :- ?X:C1.<br>- all sub-classes of one class are also subclasses of the second class:<br>?X::C1 :- ?X::C2.<br>?X::C2 :- ?X::C1. |
| owl:unionOf | <owl:Class rdf:ID="Fruit"><br><owl:unionOf rdf:parseType="Collection"><br><owl:Class rdf:about="#SweetFruit" /><br><owl:Class rdf:about="#NonSweetFruit" /><br></owl:unionOf><br></owl:Class> | Fruit :=: (SweetFruit ; NonSweetFruit) |
| owl:intersectionOf | <owl:Class rdf:ID="WhiteBurgundy"><br><owl:intersectionOf rdf:parseType="Collection"><br><owl:Class rdf:about="#Burgundy" /><br><owl:Class rdf:about="#WhiteWine" /><br></owl:intersectionOf><br></owl:Class> | WhiteBurgundy :=: (Burgundy, WhiteWine) |
| owl:complementOf | <owl:Class rdf:ID="NonConsumableThing"><br><owl:complementOf rdf:resource="#ConsumableThing" /><br></owl:Class> | NonConsumableThing :=: ( _object – ConsumableThing )<br>Note: the complement is defined using the set of all objects and the set difference operator. |
| owl:disjointWith | <owl:Class rdf:ID="Female"><br><owl:disjointWith rdf:resource="#Male" /><br></owl:Class> | disjoint_classes(Male,Female).<br>check_disjoint_constraints:-<br>disjoint_classes(?C1,?C2),<br>?X:?C1,<br>?X:?C2,<br>format('[OWL2FLORA] disjointWith constraint violation: ~w disjoint with ~w', [?C1,?C2])@_prolog(format). |
| owl:oneOf | <owl:Class rdf:ID="WineColor"><br><owl:oneOf rdf:parseType="Collection"><br><owl:Thing rdf:about="#White"/><br><owl:Thing rdf:about="#Rose"/><br><owl:Thing rdf:about="#Red"/><br></owl:oneOf><br></owl:Class> | White:WineColor.<br>Rose:WineColor.<br>Red:WineColor.<br>oneOf(WineColor,[White,Rose,Red]).<br>check_oneOf_constraints:-<br>oneOf(?C,?List),<br>?X:?C,<br>not(member(?X,?List)),<br>format(2,'[OWL2FLORA] oneOf constraint: extraneous class member ~w : ~w', [?X,?C])@_prolog(format). |
| owl:allValuesFrom | Example: For all wines, all their makers are wineries.<br><owl:Class rdf:ID="Wine"><br><rdfs:subClassOf><br><owl:Restriction><br><owl:onProperty rdf:resource="#hasMaker" /><br><owl:allValuesFrom rdf:resource="#Winery"/><br></owl:Restriction><br></rdfs:subClassOf><br></owl:Class> | F-logic constraint:<br>Wine::_object[hasMaker *=> Winery].<br>and the inference rule:<br>?Y:Winery :- ?X:Wine, ?X:[hasMaker -> ?Y]. |
| owl:someValuesFrom | Example: For all wines, they have at least one maker that is a winery.<br><owl:Class rdf:ID="Wine"><br><rdfs:subClassOf><br><owl:Restriction><br><owl:onProperty rdf:resource="#hasMaker" /><br><owl:someValuesFrom<br>rdf:resource="#Winery" /><br></owl:Restriction><br></rdfs:subClassOf><br></owl:Class> | someValuesFrom(Wine,hasMaker,Winery).<br>check_someValuesFrom_constraints:-<br>someValuesFrom(?Class,?Property,?PropertyClass),<br>?O:?Class,<br>not ?O.?Property : ?PropertyClass,<br>format(2,'[OWL2FLORA] someValuesFrom constraint violation: ~w:~w and ~w.~w disjoint from ~w', [?O,?Class,?O,?Property,?PropertyClass])@_prolog(format). |
| owl:hasValue | <owl:Class rdf:ID="Burgundy"> | hasValue(Burgundy,hasSugar,Dry). |
| | <rdfs:subClassOf><br><owl:Restriction><br><owl:onProperty rdf:resource="#hasSugar" /><br><owl:hasValue rdf:resource="#Dry" /><br></owl:Restriction><br></rdfs:subClassOf><br></owl:Class> | check_hasValue_constraints:-<br>hasValue(?Class,?Property,?Value),<br>?O:?Class,<br>not(?O[?Property->?Value]),<br>format(2,'[OWL2FLORA] hasValue constraint violation…', [])@_prolog(format). |
| owl:maxCardinality | Example: A person has no more than 2 parents.<br><owl:Class rdf:ID="Person"><br><rdfs:subClassOf><br><owl:Restriction><br><owl:onProperty rdf:resource="#hasParent"/><br><owl:maxCardinality rdf:datatype="&xsd;nonNegativeInteger">2<br></owl:maxCardinality><br></owl:Restriction><br></rdfs:subClassOf><br></owl:Class> | Person[hasParent{0:2} *=> _object].<br><br>- to check the cardilality, one may use the definitions from the Flora2 module: _typecheck.<br><br>Cardinality[%_check(?Obj[hasParent=>?])]<br>@_typecheck.. |
| rdfs:domain, rdfs:range | <owl:ObjectProperty rdf:ID="locatedIn"><br><rdfs:domain rdf:resource=" #Country" /><br><rdfs:range rdf:resource="#Region" /><br></owl:ObjectProperty> | Country[locatedIn*=>Region]. |
| rdfs:subPropertyOf | <owl:ObjectProperty rdf:ID="hasColor"><br><rdfs:subPropertyOf rdf:resource="#hasWineDescriptor" /><br></owl:ObjectProperty> | ?X[hasWineDescriptor ->?Y] :- ?X[hasColor ->?Y]. |
| owl:equivalentProperty | <owl:ObjectProperty rdf:about="#hasChild"><br><owl:equivalentProperty rdf:resource="hasOffspring"/><br></owl:ObjectProperty> | ?X[hasChild ->?Y] :- ?X[hasOffspring ->?Y].<br>?X[hasOffspring ->?Y] :- ?X[hasChild ->?Y]. |
| owl:inverseOf | <owl:ObjectProperty rdf:ID="producesWine"><br><owl:inverseOf rdf:resource="#hasMaker" /><br></owl:ObjectProperty> | ?X[producesWine ->?Y] :- ?Y[hasMaker ->?X].<br>?X[hasMaker ->?Y] :- ?Y[producesWine ->?X]. |
| owl:FunctionalProperty | <owl:ObjectProperty rdf:ID="hasVintageYear"><br><rdf:type rdf:resource="&owl;FunctionalProperty" /><br></owl:ObjectProperty> | _object[hasVintageYear{1:1} *=> _object]. |
| owl:InverseFunctional Property | <owl:ObjectProperty rdf:ID="producesWine"><br><rdf:type rdf:resource="&owl;InverseFunctionalProperty" /><br><owl:inverseOf rdf:resource="#hasMaker" /><br></owl:ObjectProperty> | ?X[producesWine ->?Y] :- ?Y[hasMaker ->?X].<br>?X[hasMaker ->?Y] :- ?Y[producesWine ->?X].<br>_object[hasMaker{1:1} *=> _object]. |
| owl:TransitiveProperty | <owl:ObjectProperty rdf:ID="locatedIn"><br><rdf:type rdf:resource="&owl;TransitiveProperty" /><br></owl:ObjectProperty> | 'TransitiveProperty'(locatedIn).<br>?X[?P->?Z]:-<br>'TransitiveProperty'(?P),<br>?X[?P->?Y],<br>?Y[?P->?Z]. |
| owl:SymmetricProperty | <owl:ObjectProperty rdf:ID="adjacentRegion"><br><rdf:type rdf:resource="&owl;SymmetricProperty" /><br></owl:ObjectProperty> | 'SymmetricProperty'(adjacentRegion).<br>?X[?P->?Y]:-<br>'SymmetricProperty'(?P),<br>?Y[?P->?X]. |

ABox axioms can also be translated between OWL and Flora2. For instance, the OWL example:

<WineGrape rdf:ID="CabernetSauvignonGrape" hasColor="Red"/>
<owl:Thing rdf:about="#PinotGrape" hasColor="White">
<rdf:type rdf:resource="#WineGrape"/>
</owl:Thing>

is translated into the class memberships and property values:

CabernetSauvignonGrape:WineGrape[hasColor->'Red'].
PinotGrape:WineGrape[hasColor->'White'].

## 3 Querying and Discussion

In most reasoning tasks, only ABox queries are considered. This fact has the obvious reason that the TBox is most of the time fixed and carefully designed, while real world queries ask questions about the ABox. Examples of this kind of reasoning can range from: applications of DL in natural language analysis, to discovery and contracting of Web services. We discuss the translation of various ABox queries into the Flora2 format. First, the ground class-instance membership queries, i.e., given a class C, determine whether a given individual a is an instance of C, are easily translated into Flora2 expressions "O:C". Similarly, open class-instance membership queries, i.e, given a class C, determine all the known individuals that are instances of C, are translated into a Flora2 query: ?L= collectset{?X | ?X:C}. Second, the "all-classes" queries (i.e., given an individual a, determine all the known classes that a is an instance of) are translated into a Flora2 query: ?L=collectset{?X | a:?X }. Third, the class subsumption queries (i.e., given classes C and D, determine if C is a subclass of D) are translated into: C::D. Forth, the class hierarchy queries (i.e, given a class C return all (or mostspecific) known super-classes of C or all (or mostgeneral) known subclasses of C) are translated into: C:?X (or "mostspecific": C:?X, not(midInh(C,?X)), where the predicate midInh(+?Sub,+?Super) detects if the class ?Sub inherits the class ?Super through a middle class: midInh(?Sub,?Super):- ?Sub::?Mid, ?Mid::?Super. Fifth, the class satisfiability queries (i.e., given a class C, determine if C is consistent) are checked using the flora2owl "check_all_constraitns" predicate and the predicates from the module _typecheck.

Disjunction is only partially translated into F-logic: when a disjunction occurs on the left hand side of a subclass axiom, e.g. C1 $\cap$ C2 $\subseteq$ D it becomes disjunction in the body of the corresponding rule: "?X:D :- ?X:C1 ; ?X:C2.", which is, basically, two Horn rules: "?X:D :- ?X:C1." and "?X:D :- ?X:C2". If the disjunction appears on the right hand side of a subclass axiom cannot be transformed into Horn F-logic rules because the disjunction in the head of the rule would blow up complexity. One way to translate a rule D $\subseteq$ C1 $\cap$ C2 is to reason by cases:

?X:C1 :- ?X:D, naf(?X:C2).
?X:C2 :- ?X:D, naf(?X:C1).

The universal quantifier is easily translated if it appears on the right hand side of a subclass axiom. A universal quantifier that appears on the left hand side of a subclass axiom can also be translated by using the Lloyd-Topor transformation (i.e., an additional predicate and negation as failure). The existential Restriction cannot be translated if it appears in the subsuming set (e.g., $\exists$X C $\subseteq$ D(X)).

The OWL numbered, qualified cardinality restrictions and the Flora2 cardinality restrictions have different semantics. OWL uses open-world assumption, so a constraint saying that there can be at most 2 elements and the knowledge base has three, OWL concludes that some 2 members of that set of three constants are equal (without telling which of two are equal). In Flora and rule-based languages in general, such a situation leads to an error. That is, Flora would say that the KB is inconsistent with the constraints. Most people believe that this is the intended meaning that people want. The owl2flora translation of the cardinality constraints does not preserve the OWL semantics and changes these cardinality constraints to the Flora2 semantics. Finally, as an additional feature, we can update the knowledge base using the Transaction Logic updates available in the Flora2 system (i.e., one rule body can insert new facts in the knowledge base, such as: "insert{?X:RedWine}").

## 4 Conclusions and Future Work

We implemented a bi-directional translation between OWL and F-logic in the Flora2 system, solving the problems met in the DL based knowledge representations: known instances querying [6], defining integrity constraint checks [7], rules [8,9], and finally, language layering in Semantic Web services architectures [10,11].